# Slim-neck by GSConv: A lightweight-design for real-time detector architectures

Hulin Li[1*], Jun Li[2], Hanbing Wei[2], Zheng Liu[3], Zhenfei Zhan[2] and Qiliang Ren[1]

1 College of Traffic and Transportation, Chongqing Jiaotong University, Chongqing 400074, China;

2 School of Mechatronics and Vehicle Engineering, Chongqing Jiaotong University, Chongqing 400074, China;

3 School of Engineering, University of British Columbia, Okanagan, Kelowna BC V1V 1V7, Canada;

alan@mails.cqjtu.edu.cn (H.L.); cqleejun@cqjtu.edu.cn (J.L.); hbwei@cqjtu.edu.cn (H.W.);

zheng.liu@ubc.ca (Z.L.); zhenfeizhan@cqjtu.edu.cn (Z.Z); qlren@cqjtu.edu.cn (Q.R.)

**Abstract –** Real-time object detection is significant for industrial and research fields. On edge devices, a giant model is difficult to achieve the real-time detecting requirement and a lightweight model built from a large number of the depth-wise separable convolutional could not achieve the sufficient accuracy. We introduce a new lightweight convolutional technique, GSConv, to lighten the model but maintain the accuracy. The GSConv accomplishes an excellent trade-off between the accuracy and speed. Furthermore, we provide a design suggestion based on the GSConv, Slim-Neck (SNs), to achieve a higher computational cost-effectiveness of the real-time detectors. The effectiveness of the SNs was robustly demonstrated in over twenty sets comparative experiments. In particular, the real-time detectors of ameliorated by the SNs obtain the state-of-the-art (70.9% $AP_{50}$ for the SODA10M at a speed of ~ 100FPS on a Tesla T4) compared with the baselines. Code is available at https://github.com/alanli1997/slim-neck-by-gsconv

## 1. Introduction

Since the RCNN [1], the deep-learning-based detectors [2-9] have dominated the object detection field. However, deep neural networks (DNNs) often have a large number of calculated parameters and consume a lot of computing resources to work. In order to mitigate high computational costs, lightweight designs of DNNs are rising [10-15]. These works mostly use the depth-wise separable convolutional (DSC) to reduce the amounts of parameters and floating-point-operations (FLOPs, the number of multiply-adds). The lightweight effect of the DSC is obvious, details in Appendix, but the disadvantages of the DSC are also obvious: the channels information of the input images is separated in the calculation process. Figure 1 (a) and (b) show the calculation process of the vanilla convolutional (SC, the standard convolutional) and the DSC. This deficiency results the DSC in a much lower representing capability of features than the SC [10, 16]. Previous lightweight works propose some solutions to alleviate the inherent flaw of the DSC: MobileNets [11-13] use a large number of 1×1 dense convolutional to fuse the channels information that is computed independently; ShuffleNets [14-15] use the "channel shuffle" to enable the interaction of channels information and GhostNet [17] uses the "cheap operation" to generate more features. But, the 1×1 dense convolutional take up more computational resources instead, the effect of using "channel shuffle" or "cheap operation" still does not touch the performance of the SC.

Many lightweight models only use the DSC from the very be-ginning to end of the deep neural networks, but the flaw of the DSC is directly amplified in the backbone whether this is used for classification or detection. We note that the features generated by only shuffling the output channels of the DSC are still the "depth-wise separated", therefore, we employ a mixing strategy to make the outputs of the DSC as close as possible to the SC by shuffling the features from the SC and the DSC - GSConv. The GSConv is an exploration for cooperating the SC and DSC. For lightweight real-time detectors, the accuracy significantly improved by only using the GSConv to replace the SC. For norm detectors, we use the SC in backbone and use the GSConv in neck (Slim-neck solution, SNs) to keep the accuracy and to infer fast. The SNs minimizes the negative impact of the flaw of the DSC and to efficiently utilize the benefits.

The main contributions of this work could be summarized as follows:

1) We introduce a new lightweight convolutional technique, GSConv, to make the representational ability of the lightweight convolutional as close to the vanilla convolutional as possible, and to reduce the computational cost.

2) We provide an efficient neck design solution, the slim-neck (SNs), for the real-time detector architecture.

3) We verified the effectiveness of different widely used tricks on popular real-time detectors, which could be a reference for study in object detection.

The rest of the article is organized as follows: we give a brief overview of some relevant works and methods about the lightweight and performance improvement of detectors in Section 2, and provide the details of the study in Section 3, experiments presented in Section 4, all the work analyzed with a discussion in Section 5 and summarized with a final conclusion in Section 6.

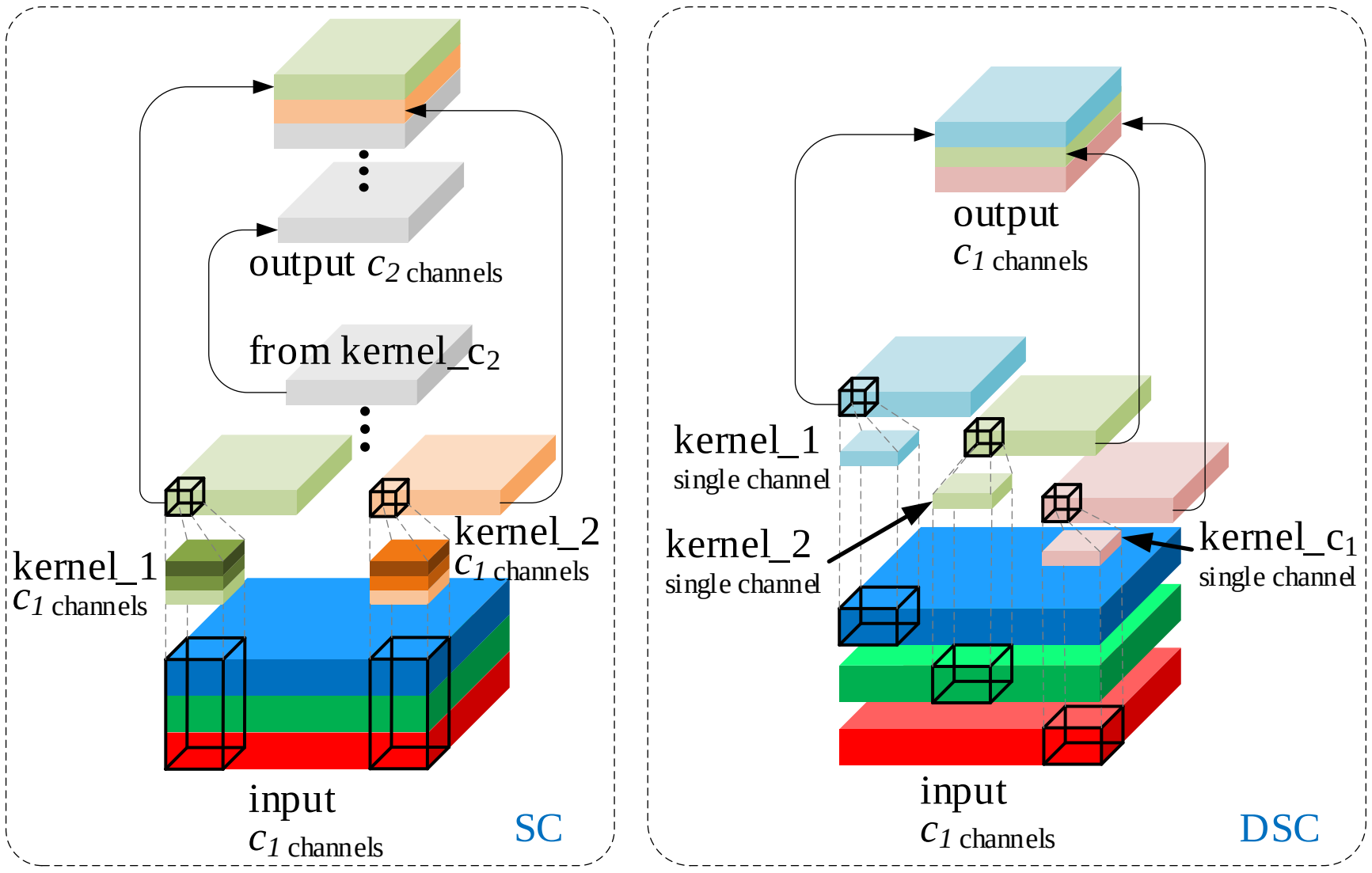

(a) (b)

Figure 1: The calculation process of the (a) SC and the (b) DSC. The SC is the channel-dense convolutional computation and the DSC is the channel-sparse.

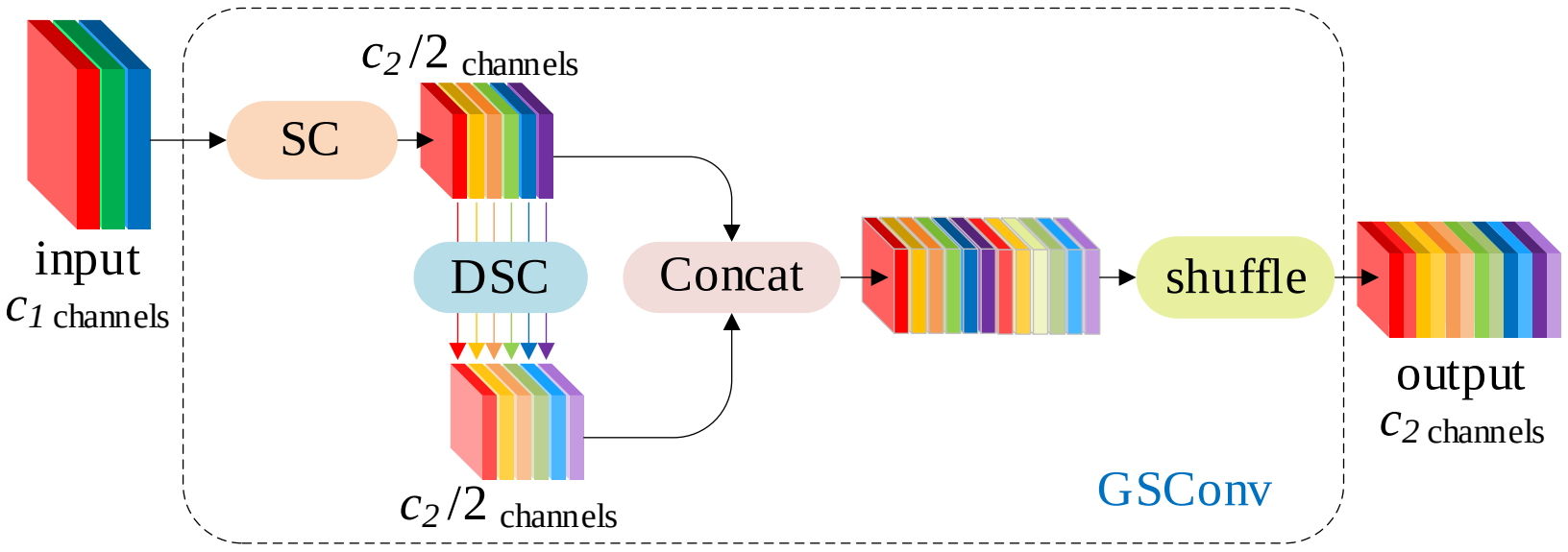

Figure 2: The structure of the GSConv. The “SC” marked in orange means the standard convolutional-2D layer with a batch normalization-2D layer and an activation layer. The “DSC” marked in blue means the depth-wise convolutional-2D layer with a batch normalization-2D layer and an activation layer.

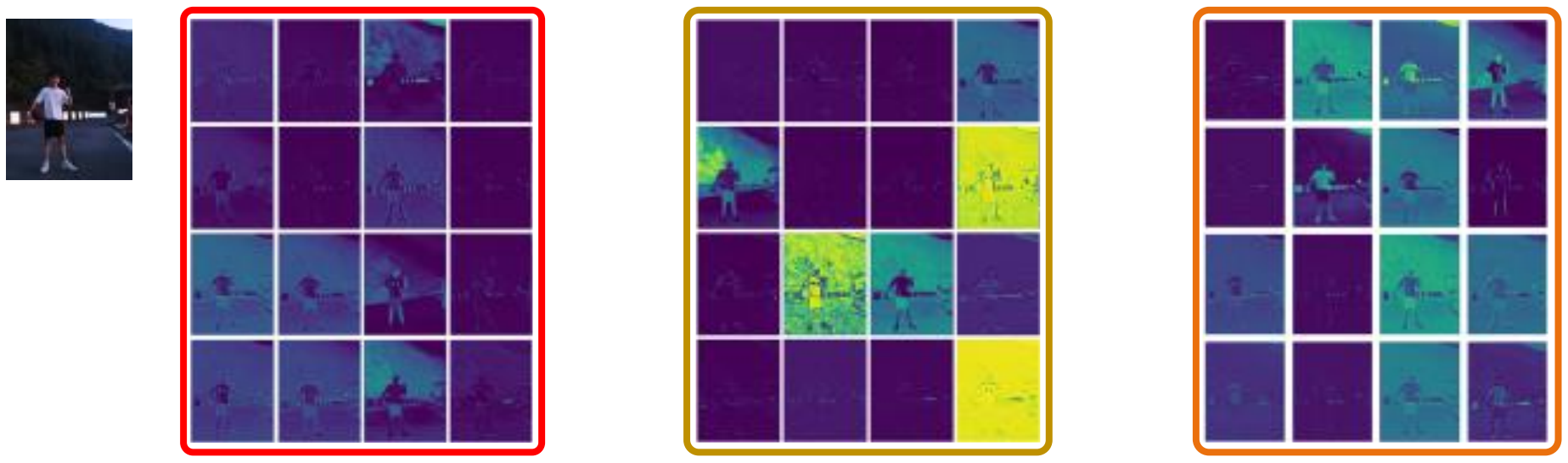

(a) features from SC (b) features from DSC (c) features from GSConv

Figure 3: Three type features of the different basic convolutional of the scale 1/4 (P2) of the Yolov5n.

## 2. Related Work

Usually, a detector based on the convolutional neural networks (CNNs) consists of three parts: a backbone, a neck and a head. The backbone for extracting features of the input, the neck for fusing features and the head for detecting objects. For the backbone, AlexNet [18] demonstrated the powerful of the features extracting of CNNs. Then, the backbone of detectors or classifiers started to be designed using the SC, such as VGG [19], ResNet [20] and DarkNet [6-7]. However, these models are unfriendly to edge computing devices because of the huge computation cost [47]. For the neck, FPN [21] improves the speed and accuracy of detectors by fusing different scale features to predict. For the head, the main difference is that the model uses anchor-based [1-3, 5-9, 25] or anchor-free [22-23] methods to accomplish localization. Moreover, the attention mechanisms, such as SPP [26], SE [27], CBAM [28] and CA [29], could improve the efficiency and performance, especially for the lightweight detectors [10-15]. We could obtain the better cost performance by a proper use of these modules, and the relevant details are described in Section 3.

## 3. **Slim-neck architecture design**

The key of this work will be described in detail in this section, including the GSConv technique, the efficient cross stage partial block (VoV-GSCSP), the free tricks for improving lightweight detectors, the proper choosing of activation functions and bounding boxes regression loss functions. We aim to build a simple and efficient Neck for the detectors to apply on the edge devices. The optimized combination of above items is the Slim-Neck optimization solution (SNs).

### 3.1 **GSConv**

In order to speed up predicting in the end, fed images almost always have to undergo a similar transformation process in the backbone: transferring the spatial information toward the channels step by step. Each time, the spaces dimension (the width and height) compression and channels dimension expansion of the features will cause a partial loss of semantic information. The channel-dense convolutional maximally preserves the hidden connections between each channel, but the channel-sparse convolution severs these connections completely. The GSConv preserves these connections as much as possible with lower time complexity, details in Appendix. As show in Figure 2, we use the shuffle to permeate the features generated by the SC (the channel-dense convolutional) into every part of the features generated by the DSC. The shuffle is a uniform mixing and allows the information from the SC to be fully mixed into the outputs of the DSC by uniformly to exchange local feature information on different channels, without bells and whistles. Figure 3 shows the visualization features from the SC, DSC and GSConv. The texture features of the GSConv are significantly more similar to the SC than the DSC to the SC.

Table 3 compares the contribution of five different basic convolutional methods (SC, DSC, ShuffleNet method, GhostNet method and GSConv) to the performance of the model. We want to complete the shuffle in GSConv in the simpler way possible and without additional FLOPs. One option is to mix features evenly by transposition operating and then reconstruct them to the original size [14], which is a way without additional FLOPs, but this may be unsupported on some edge devices. Another option is to use linear operations for the shuffle task, which costs less, but it is canonical and supported by all devices capable of performing convolution computations. The ablation study on the performance of the two shuffle methods are reported in Table 4. The advantages of the GSConv are more evident for lightweight detectors, which benefits from the enhanced nonlinear expression capability through the addition of a DSC layer and a shuffle. But if the GSConv be used at all stages of the model, the network layers will be deeper and deeper layers will aggravate the resistance to data flow and significantly increase the inference time. When these feature maps come to the neck, they have become slender enough (the channel dimension reaches the maximum, and the width and height dimensions reach the minimum) and the transformations have become moderate. Therefore, the better choice is to use the GSConv in the neck only (slim-neck + standard backbone). At this stage, using the GSConv to process the concatenated feature maps is just right: redundant repetitive information is less and compressed not needed, and the attention works better, such as the SPP and the CA.

### 3.2 **VoV-GSCSP block**

The computational cost of GSConv is about 50% of the SC, details in the Appendix, but the contribution to the model learning ability is comparable to the SC. Based on GSConv, we continue to introduce the GS bottleneck, the structure showed in Figure 4 (a). Then, we investigate the generalized methods [30-32] to enhance the learning ability of CNNs and use the one-shot aggregation strategy to design the efficient cross stage partial network (CSP) module, VoV-GSCSP, to reduce the computational complexity and inference time but maintain the accuracy. Figure 4 (b), (c) and (d) show the three design structures for the VoV-GSCSP respectively, where (b) is simple, straightforward and faster inference, and (c) and (d) have a higher reuse rate for the features. In fact, the simpler structure is used more because of the hardware friendly. Table 5 reports the results of the ablation study in detail for the three structures of VoV-GSCSP$_{1,2,3}$, and indeed, VoV-GSCSP$_1$ shows a higher performance-to-price ratio.

### 3.3 **Free improvement tricks for lightweight models**

Feature enhancement methods, attention mechanisms of vision, could improve detectors with simple structure and low computational cost. These methods reinforce certain features-weights of channel or spatial dimension. For example, the SPP concatenate four parallel branches, three maximum pooling operations (kernel size 5×5, 9×9 and 13×13) and a shortcut from the input to capture the larger spatial receptive field, and then use a point-wise convolutional layer to fuse and recalibrate spatial features-weights. A suggestion is inserting the attention modules at the end of the backbone to achieve better results, while the simple but effective SPP could be directly embedded at the entrance of the head. The reason is that the shallow networks are flooded with a large amount of low-level semantic information, resulting in the information fusion function of the attention modules being of minimal use: it is completely unnecessary to fuse features that already contain rich low-level semantic information.

### 3.4 **Loss and activation**

#### 3.4.1 **IoU loss function choosing**

The IoU [33] loss plays a great value for the detectors based on deep learning. It makes the location of prediction bounding boxes regression more accurate. As research continues to evolve, more advanced IoU loss functions have been proposed, such as GIoU [34], DIoU [35], CIoU [36] and the EIoU [37]. The CIoU-loss, Equation (1), is currently the most widely used.

$$Loss_{CIoU} = 1 - IoU + \frac{\rho^2_{(b,b^{gt})}}{d^2} + \alpha v, \quad IoU = \frac{A \bigcap B}{A \bigcup B},$$

$$\alpha = \frac{v}{(1-IoU)+v}, \quad v = \frac{4}{\pi^2}(\arctan\frac{w^{gt}}{h^{gt}} - \arctan\frac{w}{h})^2 \tag{1}$$

where the parameters "*A*" and "*B*" denote the area of ground truth bounding box and the area of prediction bounding box; the parameter "*C*" denotes the area of the minimum enclosing box of the ground truth bounding box and the prediction bounding box; the parameter "*d*" denotes the Euclidean distances of the diagonal vertices of the enclosing box; the parameter "$\rho$" denotes the Euclidean distances of the centroids of ground truth bounding box and prediction bounding box; the parameter "$\alpha$" is an indicator for trade-off, and the parameter "*v*" is an indicator to evaluate the consistency of the aspect ratio of the ground truth bounding box and prediction bounding box. However, there is an obvious problem in the CIoU to real-world:

$$\frac{\partial v}{\partial w} = \frac{8}{\pi^2}(\arctan\frac{w^{gt}}{h^{gt}} - \arctan\frac{w}{h}) \times \frac{h}{w^2+h^2},$$

$$\frac{\partial v}{\partial h} = -\frac{8}{\pi^2}(\arctan\frac{w^{gt}}{h^{gt}} - \arctan\frac{w}{h}) \times \frac{w}{w^2+h^2}$$

where the "$\partial v / \partial w$" is the gradient of the "*v*" with respect to the "*w*", the "$\partial v / \partial h$" is the gradient of the "*v*" with respect to the "*h*." According to the definition of the CIoU, if $\{(w = kw^{gt}, h = kh^{gt}) \mid k \in R^+\}$, the CIoU will degenerate into the DIoU, i.e., the relative proportion of penalty terms added in the CIoU ($\alpha v$) will not work [37]. Further, the $\partial v / \partial w$ and the $\partial v / \partial h$ have opposite signs: $\partial v / \partial w = -(h/w) \bullet (\partial v / \partial h)$. Thus, these two variables (*w* and *h*) could only be updated in the same direction, increasing or decreasing at the same time. This is not in line with practical application scenarios especially when $w < w^{gt}$ and $h < h^{gt}$ or $w > w^{gt}$ and $h > h^{gt}$. The EIoU, Equation (2), does not face this problem because it directly uses the *w* and *h* of a prediction bounding box independently for the penalty term, instead of the ratio of the *w* and *h*.

$$Loss_{EIoU} = 1 - IoU + \frac{\rho^2_{(b,b^{gt})}}{d^2} + \frac{\rho^2_{(w,w^{gt})}}{C_w^2} + \frac{\rho^2_{(h,h^{gt})}}{C_h^2} \tag{2}$$

The bounding box regression of the EIoU is more exact especially when the width of an object greatly differs the height.

### 3.4.2 **Activation function choosing**

Swish [38], Equation (3), and Mish [39], Equation (4), are generally batter than ReLU [40] for training DNNs because of smooth.

$$f(x) = x \bullet \text{sigmoid}(\beta x) \tag{3}$$

$$f(x) = x \bullet \tanh(\log(1+e^x)) \tag{4}$$

Mish performs slightly better for model accuracy than Swish on DNNs, although the two function curves are close. But compared with Swish, Mish consumes more training time because the Exponential and Logarithmic functions increase the computational cost. In Table 7 we performed adequate ablation experiments to verify the effectiveness of using different IoU-loss functions with different activation functions.

### 3.5 **The SNs for real-time detectors**

The Yolo family detectors [4-7, 44, 46] are widely used in industry because of the efficient detection. We could implement the SNs architecture by flexibility employing the GSConv, the GS bottleneck, and the VoV-GSCSP. For instance, Table 1 and 2 show the detailed comparisons between the SNs architecture and the native architecture of the Yolov4 [25] and Yolov5 [44].

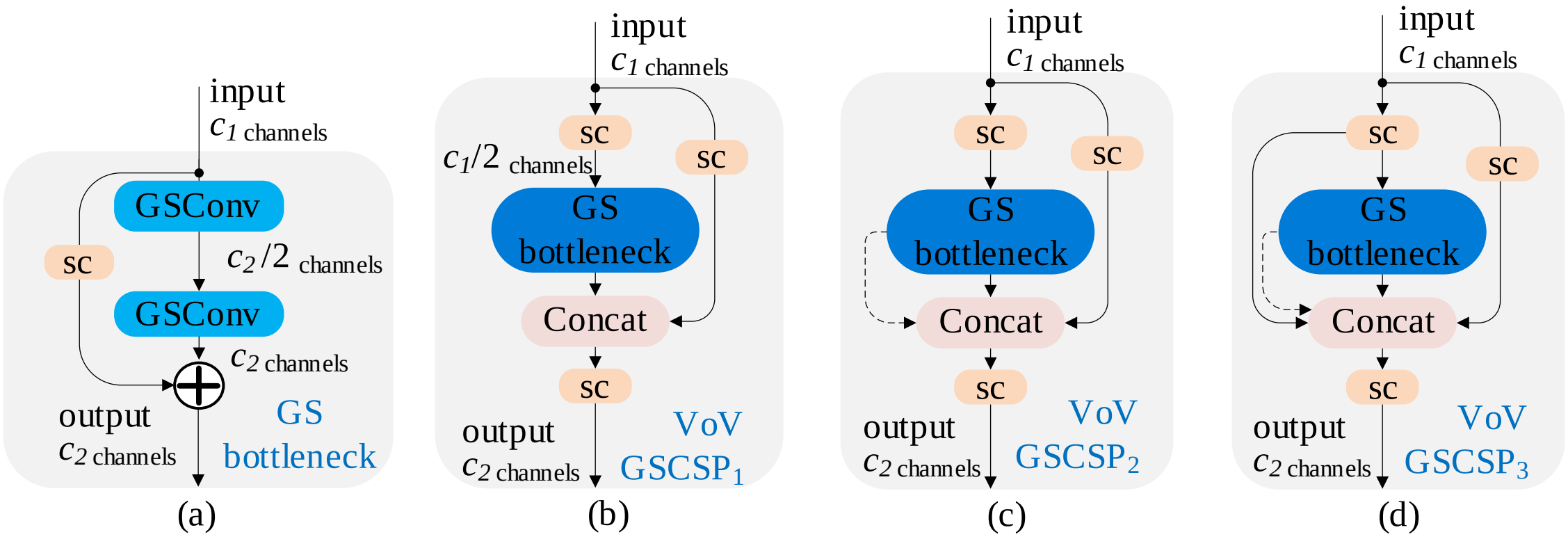

Figure 4: The structures of the (a) GS bottleneck module and the (b), (c), (d) VoV-GSCSP$_{1,2,3}$ modules.

## 4. **Experiments and Analysis**

All models are implemented by PyTorch and trained with two Nvidia Tesla T4 or one Nvidia A40 at CentOS 7 operating system. The hyper-parameters are as follows: the training steps is 33000; the optimizer is stochastic gradient descent; the batch size is 128; the linear decay learning rate scheduling strategy is adopted with initial learning rate 0.01 and final 0.001; the warm-up-steps is 1000; the momentum and weight decay are 0.937 and 0.0005. All the validation experiments are performed on a Tesla T4 or a Jetson Nano. The operation system of the Jetson Nano is Ubuntu 18.04. In tables and figures, the $AP_{50}$ is the average precision of the all categories when the accuracy evaluation IoU threshold is set to 0.5 and $AP_{95}$ is the average precision with the IoU threshold taken in steps of 0.05 from 0.5 to 0.95 and weighted average.

Table 1. Neck architecture of the SNs-Yolov4 and Yolov4

| **Neck architecture of SNs-Yolov4** | | | | **Neck architecture of Yolov4** | | |
|---|---|---|---|---|---|---|
| **Stage** | **Block** | **Channels**(I/O) | **Param.**(M) | **Block** | **Channels**(I/O) | **Param.**(M) |
| **P5** (1/32) | SPP<br>GSConv+upsample | 1024 / 512<br>512 / 256 | 1.575<br>0.069 | SPP<br>Conv+upsample | 1024 / 512<br>512 / 256 | 7.611<br>0.132 |
| **P4** (1/16) | Conv<br>VoV-GSCSP$_1$ ×2<br>GSConv+upsample | 512 / 256<br>512 / 256<br>256 / 128 | 0.130<br>0.220<br>0.018 | Conv<br>CSP<br>Conv+upsample | 512 / 256<br>512 / 256<br>256 / 128 | 0.132<br>0.986<br>0.033 |
| **P3** (1/8) | Conv<br>VoV-GSCSP$_1$ ×2<br>GSConv ×2 | 256 / 128<br>256 / 128<br>128 / 256 | 0.033<br>0.056<br>0.323 | Conv<br>CSP<br>Conv ×2 | 256 / 128<br>256 / 128<br>128 / 256 | 0.033<br>0.247<br>0.592 |
| **P3-P4** | VoV-GSCSP$_1$ ×2<br>GSConv ×2 | 512 / 256<br>256 / 512 | 0.220<br>2.361 | CSP<br>Conv ×2 | 512 / 256<br>256 / 512 | 0.986<br>2.361 |
| **P4-P5** | VoV-GSCSP$_1$ ×2<br>GSConv | 1024 / 512<br>512 / 1024 | 0.865<br>2.374 | CSP<br>Conv | 1024 / 512<br>512 / 1024 | 3.938<br>4.721 |

Table 2. Neck architecture of the SNs-Yolov5 and Yolov5

| **Neck architecture of SNs-Yolov5** | | | | **Neck architecture of Yolov5** | | |
|---|---|---|---|---|---|---|
| **Stage** | **Block** | **Channels** (I/O) | **Param.**(M) | **Block** | **Channels** (I/O) | **Param.(M)** |
| **P5** (1/32) | SPPF<br>GSConv+upsample | 1024 / 1024<br>1024 / 512 | 2.625<br>0.269 | SPPF<br>Conv+Upsample | 1024 / 1024<br>1024 / 512 | 2.625<br>0.525 |
| **P4** (1/16) | VoV-GSCSP$_1$ ×3<br>GSConv+upsample | 1024 / 512<br>512 / 256 | 0.903<br>0.069 | C3 ×3<br>Conv+Upsample | 1024 / 512<br>512 / 256 | 2.758<br>0.132 |
| **P3** (1/8) | VoV-GSCSP$_1$ ×3<br>GSConv | 512 / 256<br>256 / 256 | 0.230<br>0.298 | C3 ×3<br>Conv | 512 / 256<br>256 / 256 | 0.691<br>0.590 |
| **P3-P4** | VoV-GSCSP$_1$ ×3<br>GSConv | 512 / 512<br>512 / 512 | 0.641<br>1.187 | C3 ×3<br>Conv | 512 / 512<br>512 / 512 | 2.495<br>2.360 |
| **P4-P5** | VoV-GSCSP$_1$ ×3 | 1024 / 1024 | 2.527 | C3 ×3 | 1024 / 1024 | 9.972 |

### 4.1 **Data Sets**

We choose the WiderPerson [41] to evaluate the practical effects of the GSConv and the different attention modules. It is a pedestrian detection benchmark dataset in the wild, of which images are selected from a wide range of scenarios, no longer limited to the traffic scenario. The images collected in this dataset are all in a dense crowded environment with a large number of pedestrians and serious overlap and mutual occlusion. And, we use the VOC07+12 [42] to compare with the state-of-the-art lightweight models. Ultimately, we use the SODA10M [43] to test the real-world performance of these models in a traffic environment.

### 4.2 **Ablation Studies**

Table 3 compares five different convolution methods, the SC, DSC, ShuffleNet method, GhostNet method and the GSConv, to the accuracy of the Yolov5n on the VOC07+12 dataset. We use them at down-sample scales of 1/8, 1/16 and 1/32 in the backbone in order to verify more clearly the differences in the effectiveness. The best outcomes obtained by GSConv with the less time complexity in the end. The reason of more competitive of the GSConv is the proper cooperation of the main branch and auxiliary branch: the main branch uses the SC (channel dense) of kernel size 3×3 focusing channel hidden information, the auxiliary branch uses DSC of large kernel size 5×5 focusing spatial information, and then use a shuffle to fuse the channel and spatial information. Therefore, the GSConv could capture more spatial and channel features with less computational cost, but other convolutional methods often could not give attention to the computational cost, spatial information, and channel hidden information at the same time. Table 4 compares the performance of two shuffle approaches for the GSConv on CPU (Intel Xeon Gold 5118) and GPU (T4). In results, the transposition approach is more competitive, while linear fusion approach is a substitute when the transposition is not supported. In Table 6 we reported the experimental results of employing different attention modules by the Yolov5n. All models in the Part I use the slim-neck and slim-backbone, and models in the Part II only use the slim-neck. The "SE×3 $_{640}$" in Table 6 indicates that the SE module is used three times and SPPF once in the Yolov5n, and the training image size is 640×640.

We found that the effect of different attention modules on the number of parameters and inference time is so slight as to be almost negligible but the impact on the accuracy is significant: the "CA×3+SPPF×1" model obtains the best results with the same number of parameters on the VOC07+12 and WiderPerson. In addition, the accuracy of the slim-neck model is much better than that of the slim-backbone-neck when the inference time is very close. Further, we compare the effects of Mish and Swish activation functions, as well as CIoU and EIoU functions for accuracy and speed in Table 7. Models of using Mish with EIoU achieve the higher average accuracy, while models of using Swish achieve the faster speed (training time of using the Mish increases by about 29.26% than Swish). For contrast intuitive, we validated the effectiveness of the SNs and reported results in Table 8.

Table 3. Comparisons of the GSConv and existing methods, the SC, DSC, ShuffleNet, GhostNet (VOC07+12)

| **Basic convolutional method** | **FLOPs**(G) | **AP$_{50}$**(%) | **AP$_{95}$**(%) |
|---|---|---|---|
| SC | 4.2 | 58.9 | 35.8 |
| DSC | 3.3 | 56.9 | 34.6 |
| ShuffleNet | 3.3 | 56.5 | 34.1 |
| GhostNet | 3.8 | 59.3 | 37.0 |
| **GSConv** (ours) | 3.7 | **59.8** | **38.1** |

Table 4. Comparisons of different shuffle approaches for the GSConv (VOC07+12)

| **Shuffle option** | **AP$_{50}$**(%) | **AP$_{95}$**(%) | **FLOPs**(G) | **Latency**$_{b=1}^{CPU}$(ms) | **Latency**$_{b=1}^{GPU}$(ms) |
|---|---|---|---|---|---|
| Transposition operating | 59.8 | 38.1 | 3.7 | **37.8** | **4.3** |
| linear operating (no B.N) | 59.5 | 38.0 | 3.7 | 38.6 | 4.4 |
| linear + B.N. | 59.9 | 38.2 | 3.9 | 39.0 | 4.4 |
| **nonlinear** operating (+ ReLU) | **60.2** | **39.6** | 3.7 | 38.0 | 4.8 |

Table 5. Comparisons of three different VoV-GSCSP modules for the SNs-Yolov5n (VOC07+12)

| **Block** | **AP$_{50}$**(%) | **AP$_{95}$**(%) | **FLOPs**(G) | **Latency**$_{b=1}^{T4}$(ms) | **FPS** |
|---|---|---|---|---|---|
| VoV-GSCSP$_1$ | **61.1** | **39.3** | **4.1** | **4.5** | **168.2** |
| VoV-GSCSP$_2$ | 60.6 | 39.2 | 4.2 | 4.7 | 164.7 |
| VoV-GSCSP$_3$ | 60.8 | 36.8 | 4.3 | 4.9 | 160.0 |

Table 6. Comparisons of employment of different attention modules

| **Attention** | **Param.**(M) | **FLOPs**(G) | **P. & R.** | **AP$_{50}$**(%) | **AP$_{95}$**(%) | **Latency**$_{b=1}^{T4}$(ms) |
|---|---|---|---|---|---|---|
| **Part I. slim-backbone + slim-neck** | | | | | | |
| WiderPerson single class: pedestrian | | | | | | |
| SE×3 640 | 0.664 | 1.7 | 63.5, 78.8 | 79.9 | 46.4 | 2.2 |
| CBAM×3 640 | 0.664 | 1.7 | 63.8, 79.0 | 79.9 | 46.4 | 2.3 |
| CA×3 640 | 0.665 | 1.7 | 56.6, 82.0 | **80.4** | **47.1** | 2.2 |
| VOC07+12 20 classes | | | | | | |
| SE×3 640 | 0.685 | 1.8 | 57.3, 52.3 | 51.4 | 27.5 | 2.3 |
| CBAM×3 640 | 0.685 | 1.8 | 59.8, 51.1 | 51.2 | 27.7 | 2.4 |
| CA×3 640 | 0.685 | 1.8 | 58.5, 52.3 | **52.1** | **28.6** | 2.3 |
| **Part II. slim-neck only** | | | | | | |
| WiderPerson single class: pedestrian | | | | | | |
| SE×3 640 | 1.14 | 3.5 | 62.3, 82.9 | 82.7 | 50.6 | 1.7 |
| CBAM×3 640 | 1.14 | 3.5 | 64.1, 81.4 | 82.1 | 49.9 | 1.9 |
| CA×3 640 | 1.14 | 3.5 | 60.5, 84.4 | **83.3** | **51.5** | 1.9 |
| VOC07+12 20 classes | | | | | | |
| SE×3 640 | 1.17 | 3.5 | 64.2, 56.9 | 59.2 | 37.6 | 2.2 |
| CBAM×3 640 | 1.17 | 3.5 | 61.4, 58.2 | 58.7 | 36.8 | 2.3 |
| CA×3 640 | 1.17 | 3.5 | 64.2, 57.5 | **59.3** | **38.3** | 2.3 |

Table 7. Comparisons of the effects of Swish/Mish and CIoU/EIoU (VOC07+12)

| **Attention** | **Act. / Bbox Loss** | **AP$_{50}$**(%) | **AP$_{95}$**(%) | **Latency**$_{b=1}^{T4}$(ms) | **FPS** |
|---|---|---|---|---|---|
| CA×3, SPPF×1 640 | Swish, CIoU | 59.3 | 38.3 | 5.8 | 136.4 |
| | Swish, EIoU | 59.7 | 38.5 | **5.7** | **138.4** |
| | Mish, CIoU | 60.0 | 38.4 | 7.0 | 119.0 |
| | Mish, EIoU | **60.6** | **39.1** | 7.2 | 116.3 |

4.3 **Comparisons between the SNs detectors and baselines**

Finally, we give the performance comparisons of SNs detectors and the baselines in Table 9. The SNs detectors achieve the best accuracy with the smaller size. The combination of the SNs and the tricks makes a huge improvement on the accuracy, especially to the lightweight detectors such as the Yolov3/v4-tiny. And we put SNs-Yolo-tiny detectors on the Jetson Nano embedded device for inferring and list results in Table 10. On the low-power edge devices, we expect these models to be competent for the use cases in embedded or IoT scenarios. Figure 5 compares the speed and accuracy of five state-of-the-art detectors on the SODA10M.

Table 8. Ablation studies of three implements: the SNs, attention modules, and activation & IoU-loss functions (VOC07+12). Models marked '∗' are using the VoV-GSCSP$_1$ and marked '∗' are using the cheaper VoV-GSCSP$_1$ (a DSC layer to replace a SC layer in the GS bottleneck). Note that it is possible to obtain different FPS for the same latency due to differences in post-processing time (NMS) between different models.

| **Model** | **Param.**(M) | **FLOPs**(G) | **AP$_{50}$**(%) | **AP$_{95}$**(%) | **Latency**$^{T4}_{b=1}$(ms) | **FPS** |
|---|---|---|---|---|---|---|
| baseline (Yolov5n) | 1.79 | 4.3 | 58.0 | 35.0 | 5.8 | 136.8 |
| + slim-backbone-neck ∗ | 0.7↓**63%** | 1.8↓**58%** | 51.5↓6.5 | 27.9↓8.1 | 5.0 | 151.9 |
| + slim-backbone-neck (deep) ∗ | 1.2↓31% | 4.0↓7% | 61.1↑3.1 | 39.7↑4.7 | 6.0 | 132.7 |
| + slim-neck ∗ | 1.2↓36% | 3.5↓19% | 58.9↑0.9 | 37.1↑2.1 | **4.8** | 155.4 |
| + slim-neck ∗ | 1.7↓6% | 4.1↓5% | 61.1↑3.1 | 39.3↑4.3 | **4.8** | **168.9** |
| + slim-neck, CA×3, Mish, EIoU ∗ | 1.2↓36% | 3.5↓19% | 60.6↑2.6 | 39.1↑4.1 | 7.2 | 116.3 |
| + slim-neck, SPPF×4, CA×3, Mish, CIoU ∗ | 1.4↓23% | 3.8↓12% | 61.0↑3.0 | 39.0↑4.0 | 7.2 | 118.6 |
| + slim-neck, SPPF×4, CA×3, Mish, EIoU ∗ | 1.4↓23% | 3.8↓12% | 62.2↑4.2 | 41.5↑6.5 | 7.4 | 115.0 |
| + slim-neck, SPPF×4, CA×3, Mish, EIoU ∗ | 1.9↑7% | 4.6↑7% | **64.5**↑**6.3** | **43.9**↑**8.9** | 7.6 | 112.4 |

Table 9. Comparisons of different state-of-the-art lightweight detectors and SNs (VOC07+12)

| **Detector** | **Param.**(M) | **FLOPs**(G) | **AP$_{50}$**(%) | **AP$_{95}$**(%) | **Latency**$^{T4}_{b=1}$(ms) | **FPS** |
|---|---|---|---|---|---|---|
| **baselines:** | | | | | | |
| Yolov3-tiny [6] | 8.7 | 13.0 | 42.5 | 18.5 | 3.2 | 202.5 |
| Yolov4-tiny [7] | 6.1 | 17.6 | 45.3 | 21.3 | 3.7 | 178.6 |
| Yolov5n [44] | 1.8 | 4.3 | 58.0 | 35.0 | 5.8 | 136.8 |
| MobileNetv3-Yolov5s [13, 44] | 3.6 | 6.4 | 55.3 | 32.6 | 6.7 | 121.4 |
| ShuffleNetv2-Yolov5s [15, 44] | 5.6 | 11.6 | 56.1 | 34.5 | 5.0 | 151.3 |
| GhostNet-Yolov5s [17, 44] | 3.7 | 8.2 | 63.0 | 42.1 | 6.0 | 135.1 |
| Yolov7-tiny [46] | 6.1 | 13.2 | 65.8 | 45.4 | 6.4 | 156.3 |
| **SNs (ours):** | | | | | | |
| SNs-Yolov3-tiny | 5.9↓33% | 9.6↓23% | 56.8↑14.3 | 33.6↑15.1 | **3.0** | **219.5** |
| SNs-Yolov4-tiny | 5.7↓7% | 16.2↓8% | 61.3↑**16.0** | 37.3↑**16.0** | 3.7 | 179.1 |
| SNs-Yolov5n | 1.2↓36% | 3.5↓19% | 58.9↑0.9 | 37.1↑2.1 | 4.8 | 155.4 |
| SNs-MobileNetv3-Yolov5s | 1.8↓**50%** | 4.2↓33% | 61.9↑6.6 | 41.4↑8.8 | 8.3 | 102.9 |
| SNs-ShuffleNetv2-Yolov5s | 4.9↓13% | 7.7↓**34%** | 59.9↑3.8 | 38.7↑4.2 | 6.7 | 131.7 |
| SNs-GhostNet-Yolov5s | 3.7 - | 8.2 - | 63.6↑0.6 | 42.8↑0.7 | 6.8 | 122.8 |
| SNs-Yolov7-tiny | 5.6↓8% | 12.1↓8% | **66.3**↑0.5 | **47.7**↑2.3 | 6.4 | 156.3 |

Table 10. The performances of the SNs-Yolo-tiny on the Jetson Nano (VOC07+12)

| **Detector** | **Size** | **FLOPs**(G) | **AP$_{50}$**(%) | **AP$_{95}$**(%) | **Latency**$^{FP16}_{b=4}$(ms) | **FPS** |
|---|---|---|---|---|---|---|
| **baselines:** | | | | | | |
| Yolov3-tiny [6] | 320 | 3.3 | 49.7 | 27.9 | 31.4 | 107.9 |
| Yolov4-tiny [7] | 320 | 3.9 | 53.3 | 32.2 | 35.2 | 93.1 |
| Yolov7-tiny [46] | 320 | 3.3 | 61.9 | 42.8 | 33.1 | 102.0 |
| **SNs (ours):** | | | | | | |
| SNs-Yolov3-tiny | 320 | 2.3↓**30%** | 50.8↑1.1 | 29.4↑1.5 | 30.1 | 108.6 |
| SNs-Yolov4-tiny | 320 | 3.3↓18% | 56.8↑**3.5** | 34.5↑**2.3** | 36.0 | 91.4 |
| SNs-Yolov7-tiny | 320 | 3.0↓9% | 62.7↑0.8 | 44.2↑1.4 | 32.8 | 103.9 |

## 5. **Discussion**

### 5.1 **Real-world application of the GSConv**

At present, there are mainly two types of object detectors based on deep learning, Transformer-based and CNNs-based. The application of Transformer-based detectors faces hardship due to the bad latency and the CNNs-based detectors are still the preferred option in the industry. Researchers have proposed numerous approaches to further optimize the performance of CNNs-based models. The GSConv provides a new design option of CNNs-based vision models for field. For lightweight models, designers could directly replace the original convolutional layers by GSConv to obtain significant accuracy gains without additional operations.

But it is worth noting that the advantages of the GSConv become less obvious as the computing power of the devices grows. The GSConv is more suitable for edge computing devices because of the small computational consumption and memory footprint. In traffic application, we have tested the GSConv in some vision-based specific assisted driving systems, such as a height-limited collision monitoring and warning system for trucks and obtained promising results. In agricultural application, we believe that the GSConv is suitable for robotics or unmanned aerial vehicles (edge computing) for vision tasks such as pest and disease detection or fruit and anomaly recognition. In brief, if both cost and performance are considered, it would be a wise choice to adopt the GSConv.

### 5.2 **Extending of the GSConv**

GSConv is flexible and easily adaptable because of the initialization division of the main branch and auxiliary branch. We could add simple additional auxiliary branch for other purposes to complete specific designs and further expand the scope of application of the GSConv. For example, we know that coordinate encoding is often beneficial for detection, so adding a simple coordinate encoding auxiliary branch could optimize the detection accuracy. For another example, the visual task of low-latency frame-by-frame inference requires fewer network layers to complete fast inference (batch size is setting to one). In DNNs, however, shallow networks with small kernel size often could not capture sufficient receptive fields, while large kernel size will take huge computational cost. Fortunately, we could use the GSConv to skillfully address the difficulty by using a large kernel size DSC on the auxiliary branch. In tests, the kernel size of the DSC auxiliary branch could be extended to the 17 × 17 without causing overfitting. To be bold in exploring, we could even use the GSConv to achieve the spatial attention capability similar to the spatial pyramid: extend the auxiliary branch to three and use different kernel sizes, such as 5, 9 and 13. In fact, we have published the codes of the above potentially valid cases on the GitHub for research and trial.

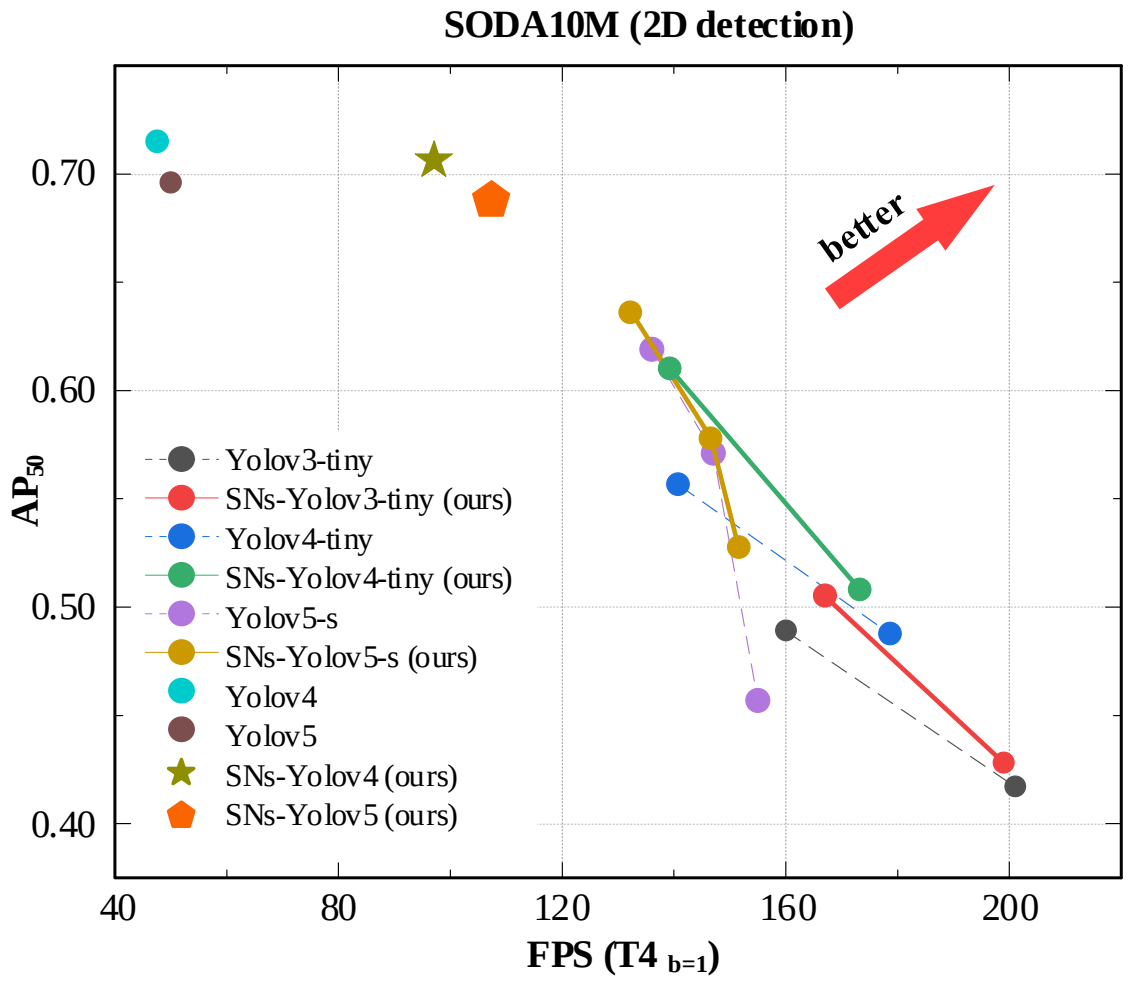

Figure 5: Comparisons of the SNs-Yolo and baselines

## 6. **Conclusion**

To sum up, in this paper we introduce a novel lightweight convolutional technique, the GSConv, to allow the depth-wise separable convolutional to achieve the close effects to the vanilla convolutional and more efficient by a simple mixing strategy of data. Further, we design the efficient cross stage partial network block, VoV-GSCSP, to replace the ordinary bottleneck to speed up inference. Moreover, we provide the slim-neck design solution (SNs) for lightweight. In experiments, the GSConv shows the better performance compared to the other lightweight convolutional methods. By the SNs, the tested detectors throw away 7.2%~32.7% of the computational cost but obtain 1.2%~35.3% improvement of the accuracy and the given results outperform state-of-the-art detectors.

**Acknowledgments**: The authors would like to thank Ms. Xinxin Liu (The Intelligent Transportation Big Data Center of the CQJTU) for her aid of the experimental GPUs providing.

## **Appendix**

### **A. The comparisons of the FLOPs and parameters amount between the DSC, SC and GSConv**

The effect of DSC is obvious in the reduction of the parameters number to detection networks. The parameters amount of a conventional convolutional layer is $C_1 \times K_1 \times K_2 \times C_2$, and the amount for a DSC is $C_1 \times K_1 \times K_2 + 1 \times 1 \times C_1 \times C_2$, where $C_1$ and $C_2$ are the channels number of the feature maps from input and output, and $K_1 \times K_2$ is the kernel size of convolutional. The calculation cost of a SC is $W \times H \times C_1 \times K_1 \times K_2 \times C_2$, and the cost of DSC is $W \times H \times C_1 \times K_1 \times K_2 + W \times H \times 1 \times 1 \times C_1 \times C_2$, where $W$ and $H$ are the width and height of the feature maps. The reason of DSC operation is cheaper than conventional convolutional could be explained by assuming the following conditions:

$$
\begin{cases}
size_{\text{input}} = W \times H \times C_1 = 320 \times 320 \times 3 \\
size_{\text{output}} = W \times H \times C_2 = 320 \times 320 \times 16 \\
size_{\text{kernel}} = K_1 \times K_2 = 3 \times 3 \\
ratio_{\text{p}} = \dfrac{C_1 \times K_1 \times K_2 + 1 \times 1 \times C_1 \times C_2}{C_1 \times K_1 \times K_2 \times C_2} = \dfrac{1}{C_2} + \dfrac{1}{K_1 \times K_2} \approx 0.174 \\
ratio_{\text{c}} = \dfrac{W \times H \times C_1 \times K_1 \times K_2 + W \times H \times 1 \times 1 \times C_1 \times C_2}{W \times H \times C_1 \times K_1 \times K_2 \times C_2} \\
\qquad = \dfrac{1}{C_2} + \dfrac{1}{K_1 \times K_2} \approx 0.174
\end{cases}
$$

where the $ratio_p$ is a ratio of the parameters amount, between DSC and conventional convolutional layer, and the $ratio_c$ is a ratio of the calculation cost of them. Obviously, the parameters amount and calculation cost of the DSC are indeed less and much lower than the SC. For the GSConv, the ratio of the computational cost of the GSConv to SC is:

$$
ratio_{\text{c}}^{*} = \frac{\frac{1}{2} \bullet W \times H \times K_1 \times K_2 \times C_2 \times (C_1 + 1)}{W \times H \times C_1 \times K_1 \times K_2 \times C_2} = \frac{C_1 + 1}{2C_1} \to \frac{1}{2}
$$

**B Comparison of remote sensing image detection**

We train the Yolov4 and the SNs-Yolov4 with an A40 on DOTA1.0 [45] (Using the same hyperparameters) to compare the ability of the two detectors to detect small objects. Figure a) and b) show the test results, and the source image is from the 23rd image of the test set of the ITCVD dataset (https://eostore.itc.utwente.nl:5001/fsdownload/zZYfgbB2X/ITCVD)

**C Comparison of SNs-Yolo-tiny and state-of-the-art lightweight detectors at night in low-light**

We test the effectiveness of our approach using a 20-second field video that was captured by a dash cam at night in low-light and show four frames from the test video as an intuitive comparison in Figure c.

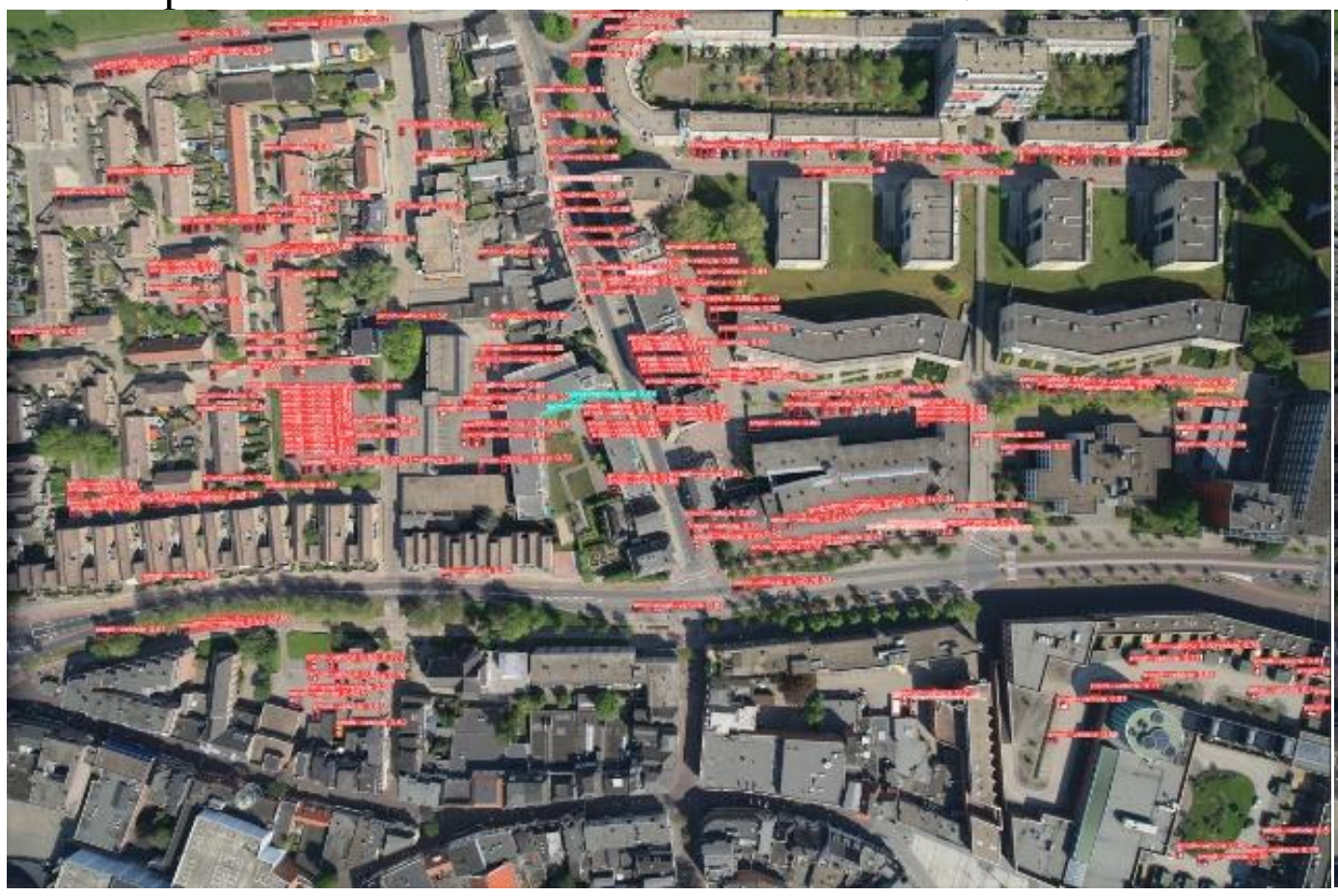

Figure a: The detection result of the SNs-Yolov4.

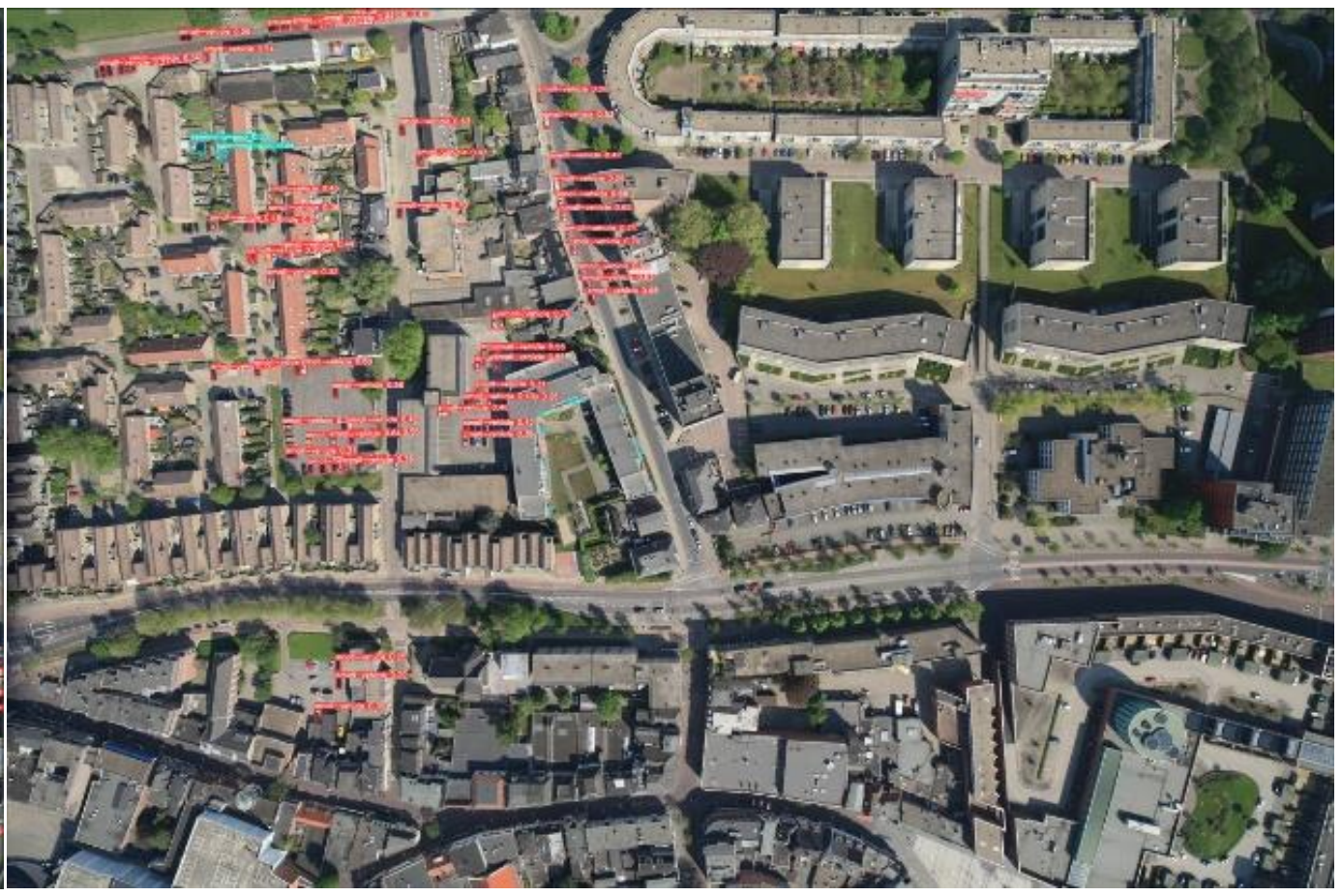

Figure b: The detection result of the original Yolov4.

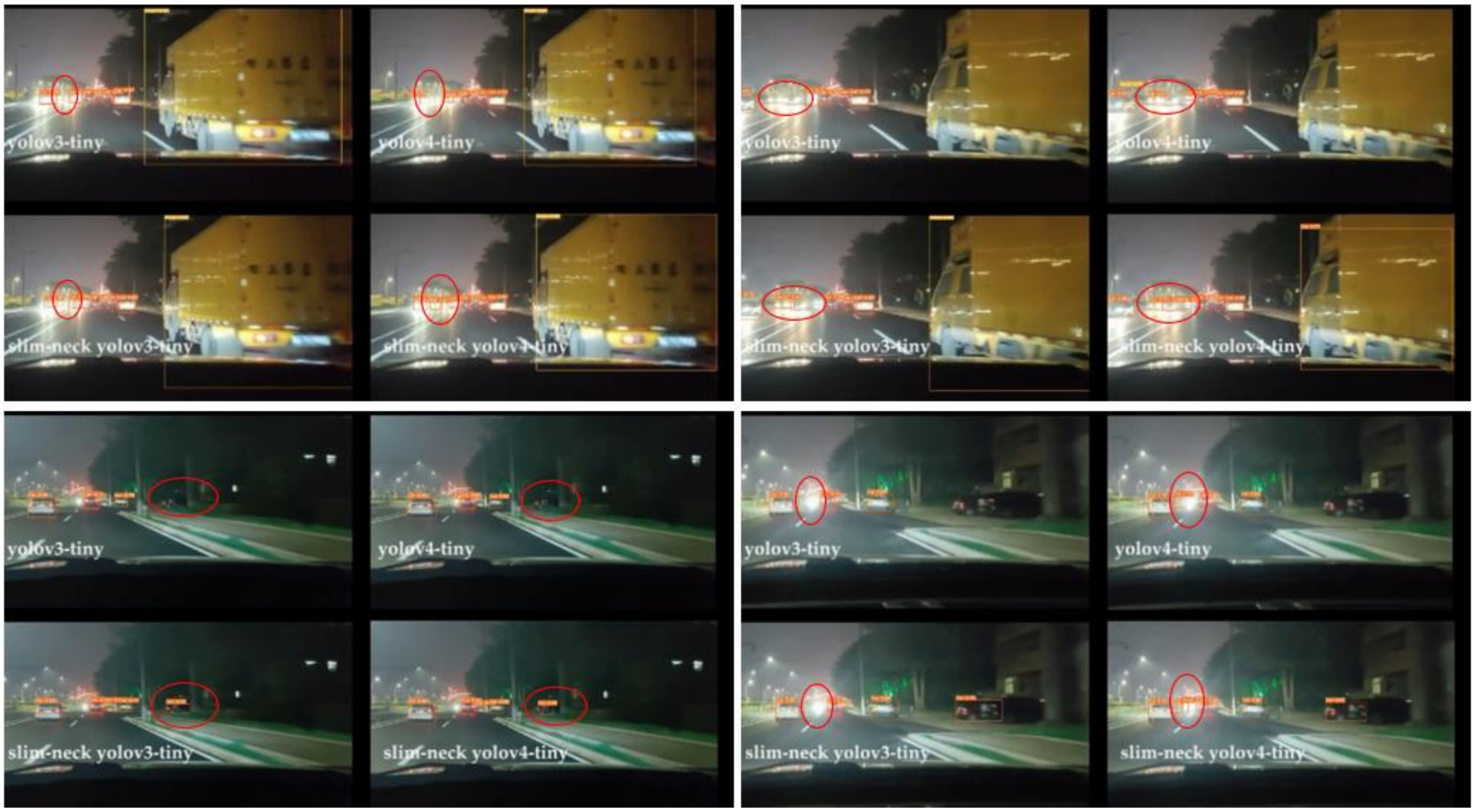

Figure c: The visual results of the SNs-Yolo in the field traffic (by a Jetson nano)